\documentclass[conference]{IEEEtran}
\IEEEoverridecommandlockouts
\usepackage{cite}
\usepackage{amsmath,amssymb,amsfonts}
\usepackage{algorithmic}
\usepackage{graphicx}
\usepackage{textcomp}
\usepackage{xcolor}
\usepackage{booktabs}
\usepackage{comment}

\def\BibTeX{{\rm B\kern-.05em{\sc i\kern-.025em b}\kern-.08em
    T\kern-.1667em\lower.7ex\hbox{E}\kern-.125emX}}
\begin{document}

\title{Author-Specific Linguistic Patterns Unveiled: \\  A Deep Learning Study on Word Class Distributions}

\author{
\IEEEauthorblockN{Patrick Krauss
\IEEEauthorblockA{
\textit{Neuroscience Lab,} \\
\textit{University Hospital Erlangen,}\\ 
\textit{CCN Group} \\
\textit{Pattern Recog. Lab.} \\
\textit{FAU Erlangen-Nürnberg}\\
Erlangen, Germany \\
patrick.krauss@fau.de}}
\and
\IEEEauthorblockN{Achim Schilling
\IEEEauthorblockA{
\textit{Neuroscience Lab,} \\
\textit{University Hospital Erlangen,}\\ 
\textit{CCN Group} \\
\textit{Pattern Recog. Lab.} \\
\textit{FAU Erlangen-Nürnberg}\\
Erlangen, Germany \\
achim.schilling@fau.de
}
}
}


\maketitle

\begin{abstract}
Deep learning methods have been increasingly applied to computational linguistics to uncover patterns in text data. This study investigates author-specific word class distributions using part-of-speech (POS) tagging and bigram analysis. By leveraging deep neural networks, we classify literary authors based on POS tag vectors and bigram frequency matrices derived from their works. We employ fully connected and convolutional neural network architectures to explore the efficacy of unigram and bigram-based representations. Our results demonstrate that while unigram features achieve moderate classification accuracy, bigram-based models significantly improve performance, suggesting that sequential word class patterns are more distinctive of authorial style. Multi-dimensional scaling (MDS) visualizations reveal meaningful clustering of authors' works, supporting the hypothesis that stylistic nuances can be captured through computational methods. These findings highlight the potential of deep learning and linguistic feature analysis for author profiling and literary studies.
\end{abstract}


\section*{Introduction}

Understanding the stylistic nuances of written language has been a long-standing goal in computational linguistics and natural language processing (NLP). One critical area of research focuses on identifying author-specific features that differentiate writing styles. Such features not only hold significance in forensic linguistics and authorship attribution \cite{stamatatos2009survey} but also offer insights into the underlying cognitive and linguistic processes that shape individual writing styles \cite{pennebaker1999linguistic}.

Deep learning has revolutionized various domains of NLP by enabling models to learn intricate patterns and relationships in text data \cite{young2018recent}. Techniques such as part-of-speech (POS) tagging and n-gram analysis have traditionally been used to study linguistic patterns \cite{jurafsky2014speech}. However, the integration of deep learning with these methods opens new avenues for exploring stylistic distinctions at a granular level. POS tags, which represent grammatical categories of words, capture structural information about language. By extending this analysis to bigram frequency distributions, sequential relationships between word classes can be examined, providing a richer representation of linguistic patterns.

In this study, we investigate author-specific linguistic patterns by combining traditional linguistic features with state-of-the-art deep learning methods. Specifically, we classify authors based on unigram (POS tags) and bigram frequency distributions extracted from their works. Using fully connected and convolutional neural networks (CNNs), we analyze how well these features differentiate authors and provide interpretable visualizations using multi-dimensional scaling (MDS) \cite{kruskal1978multidimensional}.

This paper presents a deep learning-based framework for author classification that leverages part-of-speech (POS) tags and bigram frequency distributions to capture distinctive linguistic patterns. Additionally, it evaluates the effectiveness of fully connected and convolutional neural network architectures for this task \cite{chollet2018deep}, providing a comprehensive comparison of their performance. Furthermore, the study offers novel insights into the uniqueness of linguistic patterns across authors, supported by multi-dimensional scaling (MDS) visualizations that enhance interpretability and understanding of the results.

\section*{Methods}

\subsection*{Software Resources}
The complete software for data evaluation and machine learning was written in Python 3.6. Thus, POS-tagging was performed by the usage of the German model in spaCy \cite{srinivasa2018natural, explosion2017spacy} with Python wrapper functions. Mathematical operations were performed by the usage of Numpy \cite{walt2011numpy} and for MDS projections scikitlearn \cite{pedregosa2011scikit} was used. For all visualizations Matplotlib \cite{hunter2007matplotlib} library was applied. The deep neural networks were implemented in Keras \cite{chollet2018deep} with tensorflow \cite{shukla2018machine} backend. All simulations were run on a standard desktop PC equipped with a NVidia GTX 1080 GPU.

\subsection*{Multi-dimensional Scaling (MDS)}

This technique was used to reduce the dimensionality of the hidden layer activations, preserving the pairwise distances between points as much as possible in the lower-dimensional space. In particular, MDS is an efficient embedding technique to visualize high-dimensional point clouds by projecting them onto a 2-dimensional plane. Furthermore, MDS has the decisive advantage that it is parameter-free and all mutual distances of the points are preserved, thereby conserving both the global and local structure of the underlying data \cite{torgerson1952multidimensional, kruskal1964nonmetric,kruskal1978multidimensional,cox2008multidimensional, metzner2021sleep, metzner2023extracting, metzner2022classification}. 

When interpreting patterns as points in high-dimensional space and dissimilarities between patterns as distances between corresponding points, MDS is an elegant method to visualize high-dimensional data. By color-coding each projected data point of a data set according to its label, the representation of the data can be visualized as a set of point clusters. For instance, MDS has already been applied to visualize for instance word class distributions of different linguistic corpora \cite{schilling2021analysis}, hidden layer representations (embeddings) of artificial neural networks \cite{schilling2021quantifying, krauss2021analysis}, structure and dynamics of highly recurrent neural networks \cite{krauss2019analysis, krauss2019recurrence, krauss2019weight, metzner2023quantifying}, or brain activity patterns assessed during e.g. pure tone or speech perception \cite{krauss2018statistical, schilling2021analysis}, or even during sleep \cite{krauss2018analysis, traxdorf2019microstructure, metzner2022classification, metzner2023extracting}. 
In all these cases the apparent compactness and mutual overlap of the point clusters permits a qualitative assessment of how well the different classes separate.

\subsection*{Data Set}

The dataset used in this study comprises a collection of 193 literary works authored by 76 different writers. These texts span a variety of genres and time periods, providing a diverse linguistic corpus for analysis. The dataset was curated to ensure that each author contributed at least one complete text, with some authors represented by multiple works.

To prepare the data for analysis, the texts were tokenized and annotated with part-of-speech (POS) tags using the German language model in spaCy \cite{srinivasa2018natural, explosion2017spacy}. Two types of feature representations were extracted for each text: unigram frequency vectors (POS tags) and bigram frequency matrices (sequential word class combinations). These features serve as the basis for the deep learning models employed in this study.

Preprocessing steps included removing punctuation, converting all text to lowercase, and excluding texts with fewer than 1,000 words to ensure statistical reliability. The final dataset was split into training, validation, and test sets using an 80/10/10 split, stratified by author, to maintain a balanced representation of linguistic styles across subsets. This ensures that the evaluation of the models reflects their ability to generalize to unseen data while preserving author-specific characteristics.

\subsection*{Neural network architecture}

To classify authors based on linguistic patterns, two deep neural network architectures were developed and implemented: a fully connected network for POS-tag vectors and a convolutional neural network (CNN) for bigram frequency matrices. Each architecture was specifically tailored to the input data's characteristics to optimize performance and ensure interpretable results.

\subsubsection*{Fully Connected Network (POS-Tag Vectors)}
The fully connected network processes unigram frequency vectors derived from POS tags. The architecture consists of a series of dense layers with dropout layers added to prevent overfitting (Table \ref{tab:DNN_POS}). The network begins with an input layer of 11 dimensions (corresponding to the 11 POS tags) and progresses through multiple dense layers with ReLU activations. The final output layer uses a softmax activation function to produce an 8-class probability distribution, corresponding to the target authors. Dropout rates between 0.5 and 0.6 were applied to ensure robust generalization during training.

\begin{table}[ht!]
    \centering
    \begin{tabular}{c||c c c c}
         Layer& Type & input-output-dim & activation & characteristics\\
         \hline
         \hline
1 & Dense & 11; 20 & relu & ~\\
2 & Dropout & 20; 20 & ~ & dropout: 0.6\\
3 & Dense & 20; 18 & relu & ~\\
4 & Dropout & 18; 18 & ~ & dropout: 0.5\\
5 & Dense & 18; 16 & relu & ~\\
6 & Dropout & 16; 16 & ~ & dropout: 0.5\\
7 & Dense & 16; 15 & relu & ~\\
8 & Dropout & 15; 15 & ~ & dropout: 0.5\\
9 & Dense & 15; 8 & softmax & ~\\
        
    \end{tabular}
    \caption{\textbf{Exact parameters of the feed foward network for author classification using POS-tag vectors}}
    \label{tab:DNN_POS}
\end{table}

\subsubsection*{Convolutional Neural Network (Bigram Frequency Matrices)}
The CNN is designed to capture sequential dependencies and spatial patterns within the bigram frequency matrices. This network incorporates convolutional layers followed by max-pooling layers to extract local features (Table \ref{tab:DNN_Bigram}). The extracted features are then flattened and passed through dense layers for further processing. Dropout layers with a rate of 0.3 were included to reduce the risk of overfitting. The output layer, like the fully connected network, utilizes a softmax activation function to classify authors into one of eight categories.

\begin{table}[ht!]
    \centering
    \begin{tabular}{c||c c c c}
         Layer& Type & input-output-dim & activation & characteristics\\
         \hline
         \hline
1 & Convolution 2D & 11; 11; 1 & relu & ~\\
2 & MaxPooling & ~ & ~ & ~\\
3 & Convolutio 2D & ~& relu & ~\\
4 & MaxPooling & ~ & ~ & ~\\
5 & Faltten & ~ & relu & ~\\
6 & Dense & x; 30 & relu & ~\\
7 & Droput & 30; 30 & ~ & dropout: 0.3\\
8 & Dense & 30; 18 & relu & ~\\
9 & Droput & 18; 18 & ~ & dropout: 0.3\\
10 & Dense & 18; 16 & relu & ~\\
11 & Droput & 16; 16 & ~ & dropout: 0.3\\
12 & Dense & 16; 15 & relu & ~\\
13 & Droput & 15; 15 & ~ & dropout: 0.3\\
14 & Dense & 15; 8 & softmax & ~\\
        
    \end{tabular}
    \caption{\textbf{Exact parameters of the feed-forward network for author classification using bigram-vectors}}
    \label{tab:DNN_Bigram}
\end{table}

Both networks were implemented using the Keras library with a TensorFlow backend. Training was performed using categorical cross-entropy as the loss function and the Adam optimizer for efficient convergence. The networks were trained on a standard desktop equipped with an NVIDIA GTX 1080 GPU, ensuring rapid iteration and optimization.

\section*{Results}

\subsection*{Author-Specific Word Class Distributions}

\begin{figure}[ht!]
	\centering
	\includegraphics[width=0.8\linewidth]{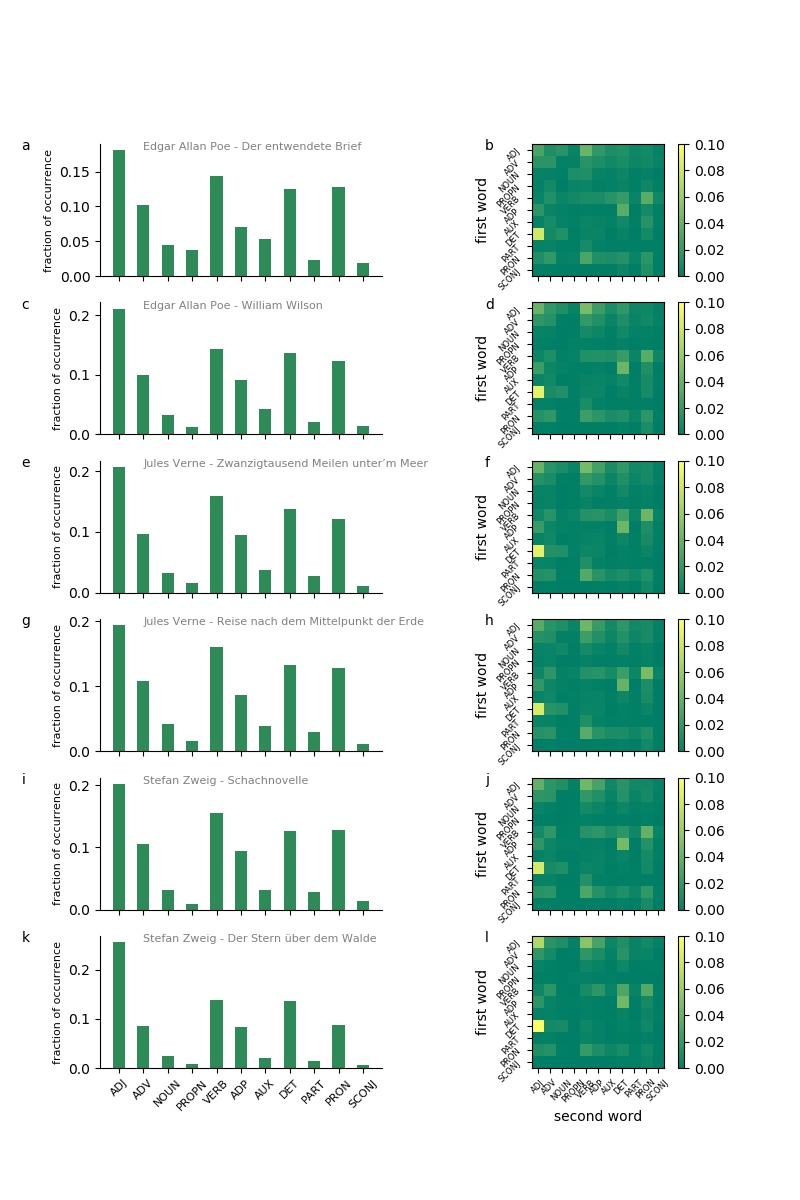}
	\caption{\textbf{Histograms of POS tags and bigram matrices}\newline The plot shows the frequency of occurrence of several word classes (POS tags, a, c, e, g, i, k) normed by the number of words and the normed frequency of occurrence of word class combinations of consecutive words (bigrams, b, d, f, h, j, l). The results for two books of three authors are shown (Edgar Allan Poe, Jules Verne, Stefan Zweig) .} 
	\label{Histograms}
\end{figure}

Figure \ref{Histograms} illustrates the normalized frequency distributions of part-of-speech (POS) tags and bigram matrices for two works each by Edgar Allan Poe, Jules Verne, and Stefan Zweig. The histograms reveal distinct patterns in the usage of individual word classes (POS tags) and their sequential combinations (bigrams), suggesting that authors exhibit unique stylistic tendencies in their linguistic choices. For instance, differences in the frequency of POS tags such as determiners (a), verbs (g), and adjectives (i) are evident across the authors, reflecting variations in sentence structure and lexical preferences. Similarly, the bigram distributions (e.g., b, d, f) highlight unique combinations of word classes that further distinguish the authors’ styles. These findings underscore the potential of POS tags and bigram matrices as effective features for capturing authorial style, forming the basis for subsequent neural network-based classification tasks.

\subsection*{Multi-Dimensional Scaling of Linguistic Patterns}

\begin{figure}[ht!]
	\centering
	\includegraphics[width=1.0\linewidth]{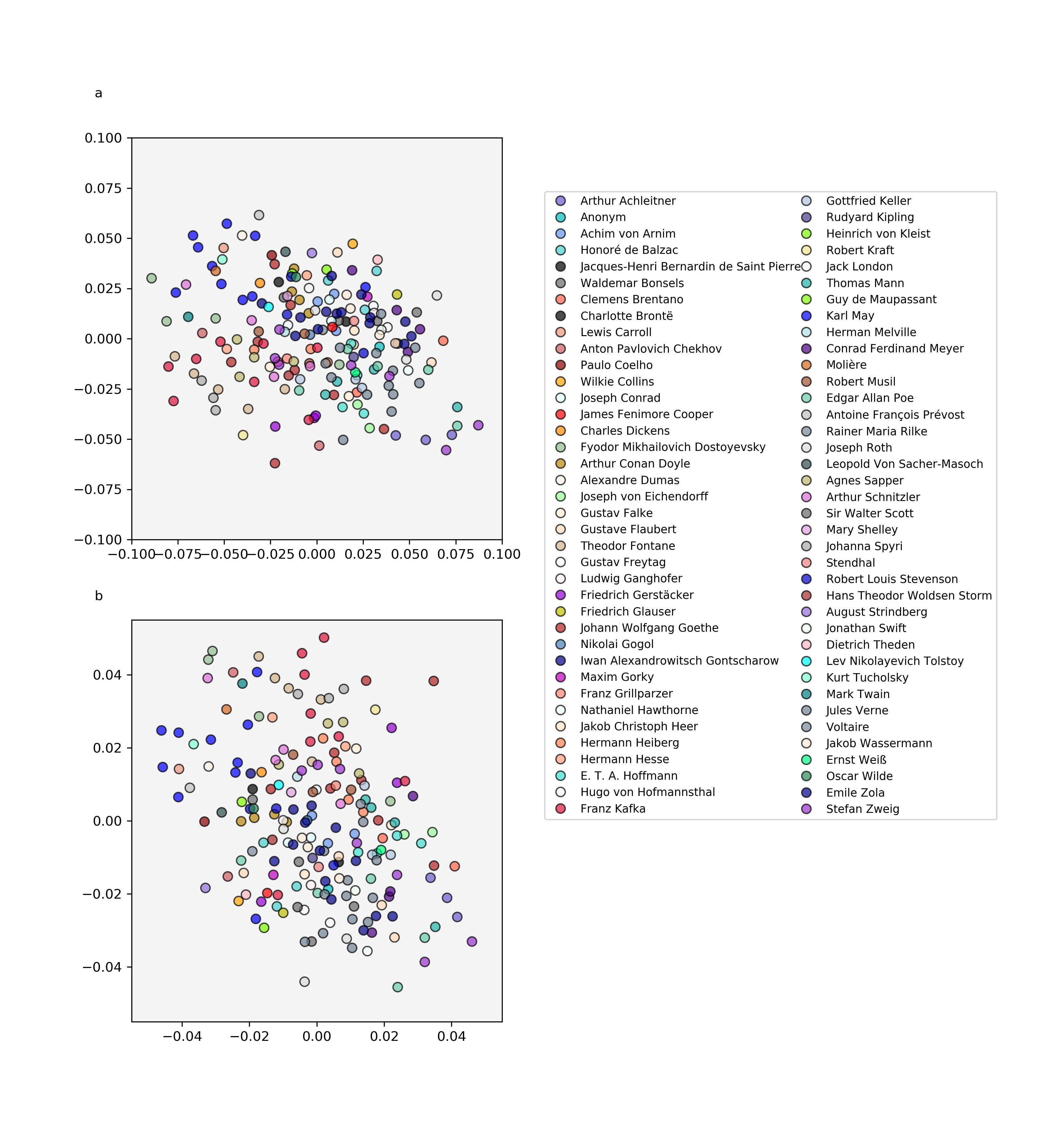}
	\caption{\textbf{MDS plot of the unigram (a) and bigram (b) frequency vectors of 193 books from 76 different authors}\newline a: Multi-Dimensional scaling of 193 POS-tag-vectors referring to the literary works of 76 others. b: Analog anlysis to a for bigram matrices. Bigram matrices 11x11 were flattened before the MDS procedure.} 
	\label{MDS_All}
\end{figure}

Figure \ref{MDS_All} presents multi-dimensional scaling (MDS) plots that visualize the unigram (POS-tag vectors, panel a) and bigram (flattened bigram matrices, panel b) frequency distributions for 193 books authored by 76 different writers. The MDS projections reveal clear clustering patterns, indicating that linguistic features such as POS tags and bigram combinations encapsulate distinctive stylistic traits associated with individual authors. Panel (a) demonstrates the grouping of texts based on unigram distributions, while panel (b) highlights even more pronounced separations using bigram features, reflecting the added complexity and sequential information captured in these representations. These clusters provide compelling evidence that authors exhibit consistent linguistic styles across their works, which can be effectively visualized and quantified through MDS techniques. This analysis lays the foundation for leveraging these features in classification models.

\subsection*{Clustering Analysis of Frequent Authors}

\begin{figure}[ht!]
	\centering
	\includegraphics[width=1.0\linewidth]{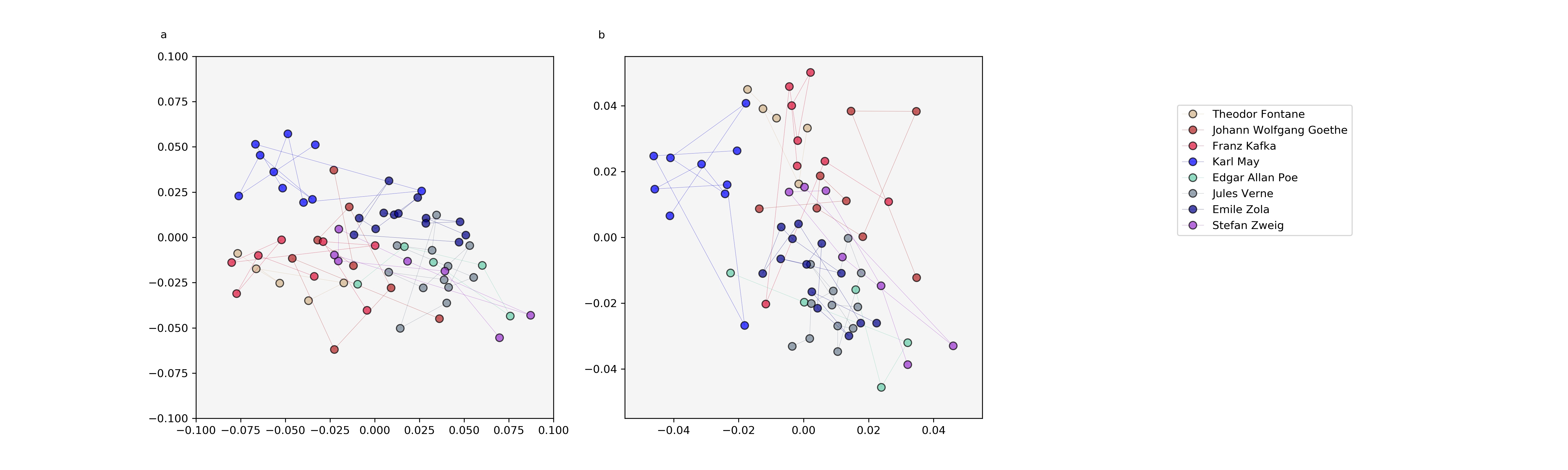}
	\caption{\textbf{MDS analysis for all authors which occur more than 5 times in the data set}\newline Note that no new MDS was performed (same as in Fig. \ref{MDS_All}) a: POS-tag vectors, b: Bigram-matrices} 
	\label{MDS_4occ}
\end{figure}

Figure \ref{MDS_4occ} focuses on the multi-dimensional scaling (MDS) projections for authors with at least five works in the dataset, using the same analysis as Figure \ref{MDS_All}. The POS-tag vector plot (panel a) and the bigram matrix plot (panel b) both reveal distinct and tighter clustering for these frequent authors, suggesting greater consistency in linguistic style across their works. The separation between clusters remains more pronounced in the bigram matrix analysis (panel b), reflecting the utility of sequential word class patterns for distinguishing authors. These results emphasize that frequent authors exhibit robust stylistic patterns that can be captured effectively through both unigram and bigram features, with the latter providing higher discriminatory power. This subset analysis highlights the reliability of these features for author classification, particularly in scenarios with a sufficient volume of textual data per author.

\subsection*{Cluster Stability and Variability in Linguistic Patterns}

\begin{figure}[ht!]
	\centering
	\includegraphics[width=1.0\linewidth]{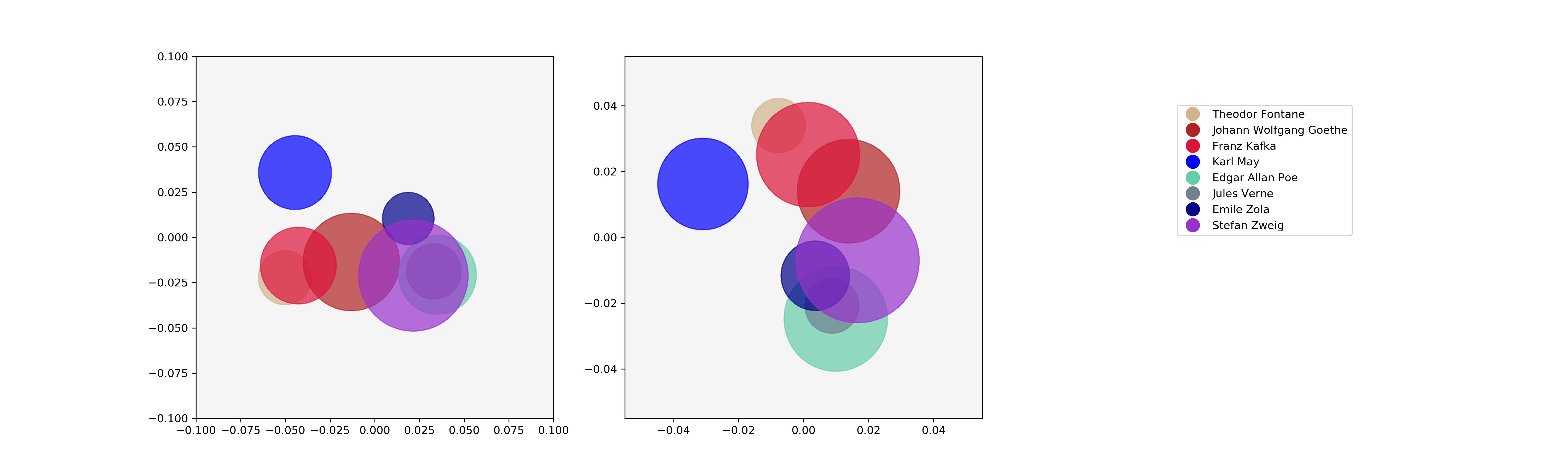}
	\caption{\textbf{Center of mass and deviation from the center of mass for MDS plots}\newline a: POS-tag vectors, b: Bigram-matrices; Each cluster of the 8 different authors is represented by the center of mass of the cluster (focus of the circle) and the average standard deviation in 2D (radius of the circle).} 
	\label{MDS_4occ_circ}
\end{figure}

Figure \ref{MDS_4occ_circ} illustrates the center of mass and deviation for clusters in the multi-dimensional scaling (MDS) projections of linguistic features for eight authors with at least five works in the dataset. In both the POS-tag vector plot (panel a) and the bigram matrix plot (panel b), each cluster is represented by its center of mass (circle center) and the average standard deviation (circle radius). The visualization reveals consistent clustering across both feature types, with tighter clusters observed for bigram matrices (panel b), indicating lower variability in sequential word class patterns compared to individual POS-tag frequencies. The varying sizes of the circles reflect differences in stylistic variability among authors, suggesting that some authors maintain a more consistent linguistic style across their works. This analysis underscores the effectiveness of bigram features in capturing stable stylistic patterns and highlights the importance of variability measures in authorial style analysis.

\subsection*{Author Classification Using POS-Tag Vectors}

\begin{figure}[ht!]
	\centering
	\includegraphics[width=1.0\linewidth]{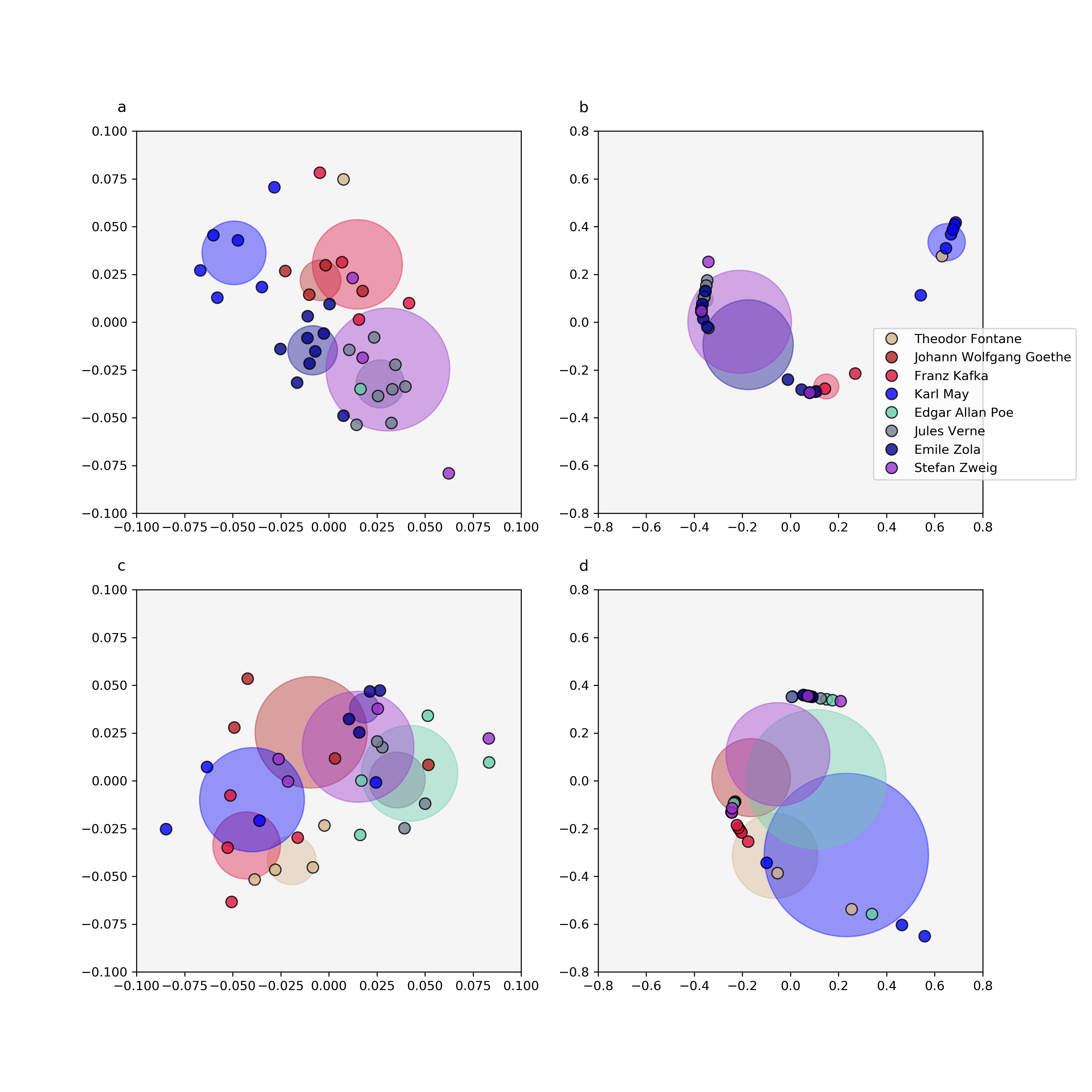}
	\caption{\textbf{Deep learning with POS-tag vectors}\newline An simple fully-connected network was trained on author classification. a: Training data set projected using MDS, b: The output of the last layer i.e. embeddings  (softmax layer) projected using MDS. c: Test data set projected with MDS, d: Embeddings of test data set. Author classification by the simple usage of POS-tag vectors leads to test accuracy smaller than 0.5 (training accuracy: 0.61, test accuracy: 0.44).} 
	\label{Unigrams_DL}
\end{figure}

Figure \ref{Unigrams_DL} demonstrates the performance of a simple fully connected neural network trained to classify authors based on POS-tag vectors. The multi-dimensional scaling (MDS) plots visualize the training dataset (panel a) and the corresponding embeddings from the final layer of the network (panel b). Similarly, the test dataset (panel c) and its embeddings (panel d) are shown. The clustering observed in the embeddings indicates that the network captures some degree of separation among authors, although the classification performance is modest. The training accuracy reached 0.61, while the test accuracy was 0.44. Importantly, the test accuracy is well above the chance level of 0.125 (1/8, based on the number of classes), demonstrating the feasibility of using POS-tag vectors for author classification. These results highlight the potential of POS-tag features to capture stylistic distinctions, even though they are limited in capturing more complex sequential patterns that could further improve classification performance.

\subsection*{Author Classification Using Bigram Matrices}

\begin{figure}[ht!]
	\centering
	\includegraphics[width=1.0\linewidth]{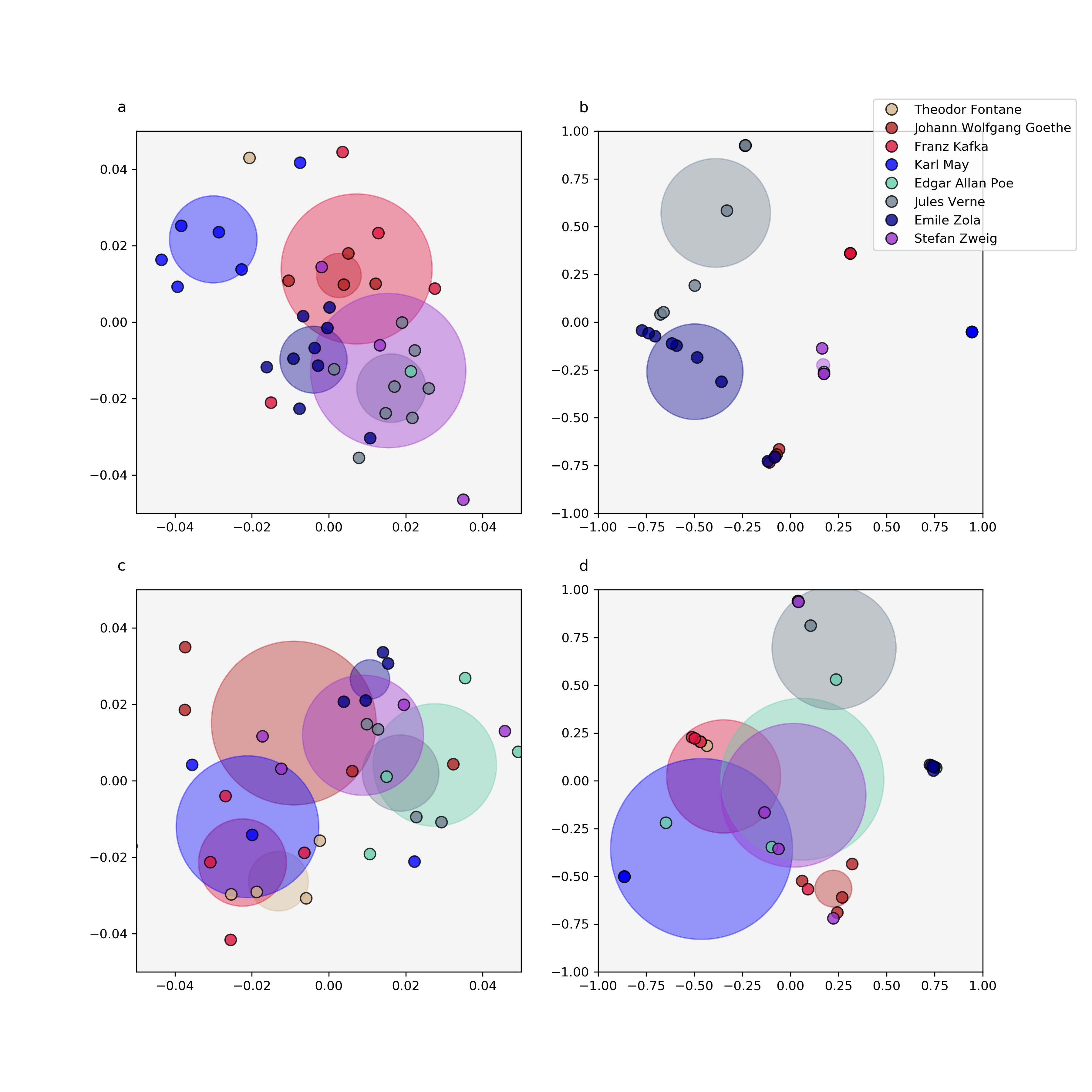}
	\caption{\textbf{Deep learning with bigram-matrices}\newline A convolutional neural network was trained on author classification based on bigram matrices (11x11). a, c: Training resp. test data set projected using MDS; b, d: Embeddings of training and test data set (output of softmax layer) projected using MDS (training accuracy: 0.81, test accuracy: 0.59). } 
	\label{Bigrams_DL}
\end{figure}

Figure \ref{Bigrams_DL} showcases the performance of a convolutional neural network (CNN) trained to classify authors based on bigram frequency matrices (11x11). The multi-dimensional scaling (MDS) plots display the training dataset (panel a) and the embeddings from the final layer of the network (panel b), as well as the test dataset (panel c) and its corresponding embeddings (panel d). The CNN achieves a training accuracy of 0.81 and a test accuracy of 0.59, which is significantly higher than the chance level of 0.125 (1/8, based on the number of classes). The embeddings in both the training and test datasets exhibit tighter and more distinct clustering compared to the POS-tag vector model, indicating the superior ability of bigram features to capture sequential patterns in authorial style. These results demonstrate that bigram matrices provide a more effective representation for author classification, offering improved generalization and performance compared to unigram-based models.

\section*{Discussion}
This study investigates the potential of deep learning methods to classify authors based on linguistic patterns, specifically unigram (POS-tag vectors) and bigram frequency matrices. Our findings highlight the utility of these features for capturing authorial style and demonstrate the ability of neural networks to leverage these patterns for classification tasks. While the results are promising, several key insights and limitations merit discussion.
Key Insights

The results reveal that bigram frequency matrices are significantly more effective than unigram POS-tag vectors for author classification. This is evident from the higher test accuracy (0.59) achieved by the convolutional neural network (CNN) trained on bigram features compared to the fully connected network's performance on POS-tag vectors (test accuracy of 0.44). The enhanced performance of bigram features underscores the importance of sequential patterns in distinguishing stylistic nuances, as they capture contextual relationships between word classes that unigrams cannot.

The MDS visualizations further support these findings by illustrating more distinct and tighter clustering in the embeddings of bigram features. This suggests that the CNN effectively learns meaningful representations of the input data, facilitating better separation of authors' works. Moreover, the test accuracies for both models exceed the chance level of 0.125, demonstrating the feasibility of using linguistic features for author classification and validating the underlying principles of this approach.

\subsection*{Limitations and Challenges}

Despite these promising results, the test accuracy of the models, particularly for the POS-tag vectors, remains modest. This highlights the inherent limitations of unigram features in capturing the complexity of authorial style, as they lack sequential and contextual information. While bigram features address this to some extent, the test accuracy of 0.59 suggests room for improvement. This could stem from factors such as variability in linguistic style across an author's works, noise in the dataset, or the relatively simple neural architectures employed.

Another limitation is the reliance on a fixed set of linguistic features (POS tags and bigram matrices), which may not fully capture all stylistic dimensions. Incorporating additional features, such as syntactic structures or semantic embeddings, could provide a more holistic representation of authorial style. Furthermore, the dataset size, while sufficient for this study, may constrain the generalizability of the findings, particularly for authors with fewer works.

\subsection*{Future Directions}

Future research could explore more advanced neural architectures, such as recurrent neural networks (RNNs) or transformers, to better capture sequential dependencies and hierarchical patterns in text. Additionally, incorporating multilingual datasets and applying transfer learning could extend the applicability of this approach to a broader range of linguistic contexts.

Another promising avenue is the integration of multimodal features, such as combining linguistic patterns with metadata (e.g., publication dates, genres) to enhance classification accuracy. Expanding the dataset to include a more diverse range of authors and works could also improve the robustness and generalizability of the models.

\section*{Conclusion}

This study demonstrates the potential of deep learning for author classification using linguistic features, with bigram matrices emerging as a particularly effective representation. While there is room for improvement, the findings provide a solid foundation for future work aimed at developing more sophisticated models and feature representations. By advancing our ability to analyze and classify authorship, this research contributes to the broader fields of computational linguistics, authorship attribution, and digital humanities.

\section*{Additional Information}

\subsection*{Data availability statement} 
All data will be made available upon request.

\subsection*{Code availability statement} 
All codes will be made available upon request.

\section*{Author contributions}
AS and PK contributed equally to this work.

\section*{Competing interests}
The authors declare no competing financial interests.

\section*{Acknowledgements}

This work was funded by the Deutsche Forschungsgemeinschaft (DFG, German Research Foundation): KR\,5148/3-1 (project number 510395418), KR\,5148/5-1 (project number 542747151), and GRK\,2839 (project number 468527017) to PK, and grant SCHI\,1482/3-1 (project number 451810794) to AS. 
We thank Alexandra Zankl for technical assistance.



\end{document}